# A Generic Framework for Assessing the Performance Bounds of Image Feature Detectors

**Shoaib Ehsan [1,\*], Adrian F. Clark [1], Ales Leonardis [2], Naveed ur Rehman [3] and Klaus D. McDonald-Maier [1]**

[1]  School of Computer Science and Electronic Engineering, University of Essex, Colchester CO4 3SQ United Kingdom; E-Mails: sehsan@essex.ac.uk; alien@essex.ac.uk; kdm@essex.ac.uk

[2]  Department of Electrical Engineering, University of Birmingham, Birmingham, United Kingdom; E-Mails: a.leonardis@cs.bham.ac.uk

[3]  Department of Electrical Engineering, COMSATS Institute of Information Technology, Islamabad, Pakistan; E-Mails: naveed.rehman@comsats.edu.pk

\*  Author to whom correspondence should be addressed; E-Mail: sehsan@essex.ac.uk ;
   Tel.: +44-1206-874376; Fax: +44-1206-872788.

**Abstract:** Since local feature detection has been one of the most active research areas in computer vision during the last decade, a large number of detectors have been proposed. The interest in feature-based applications continues to grow and has thus rendered the task of characterizing the performance of various feature detection methods an important issue in vision research. Inspired by the good practices of electronic system design, a generic framework based on the repeatability measure is presented in this paper that allows assessment of the upper and lower bounds of detector performance and finds statistically significant performance differences between detectors as a function of image transformation amount by introducing a new variant of McNemar's test in an effort to design more reliable and effective vision systems. The proposed framework is then employed to establish operating and guarantee regions for several state-of-the art detectors and to identify their statistical performance differences for three specific image transformations: JPEG compression, uniform light changes and blurring. The results are obtained using a newly acquired, large image database (20482 images) with 539 different scenes. These results provide new insights into the behavior of detectors and are also useful from the vision systems design perspective.

## 1. Introduction

Consider designing a small power supply for one of the coldest inhabited regions on the planet – Oymyakon, a village in Russia (Siberia). Apart from some general design constraints, such as the required maximum output voltage, freezing temperatures in the area of deployment make this design task more challenging as the characteristics of most electronic components change with temperature. For example, the capacitance of a capacitor is a function of temperature. Similarly, the current-voltage characteristics of a diode are also dependent upon temperature. Thus, looking at the *datasheets* of the required electronic components for this power supply would be a logical step for finding devices that operate reliably in extremely low temperatures. Only those components would be selected which show stable operating characteristics across the required range of temperatures to ensure that the final output of the power supply would satisfy the initial design specifications.

**Figure 1.** Two sample images; the left image is the reference image whereas the right image undergoes 20% uniform decrease in illumination.

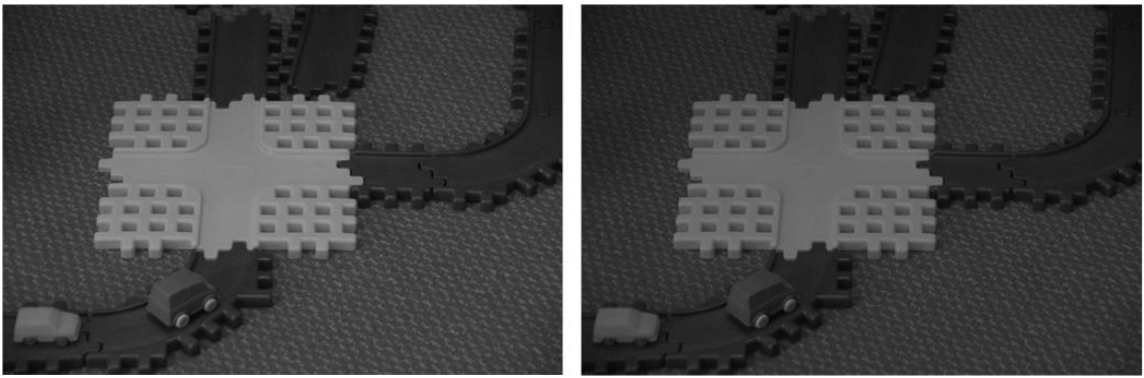

Now come back to the computer vision world and design a simple toy car tracking system with local feature detection as its primary stage while expecting only 20% uniform decrease in illumination. Looking at the repeatability results presented in [1] (which are widely considered the most comprehensive) for the Leuven dataset (which involves uniform changes in light) [2], MSER detector [3] appears to be the best option for achieving a reasonable value of repeatability (more than 60%) for this small transformation amount. Now consider two sample images shown in Figure 1 which the designed vision system would encounter when deployed in the actual environment. The first image is the reference image and the second image has undergone 20% uniform decrease in light relative to the reference. Theoretically speaking, the feature detection unit (based on MSER) of the designed vision system would achieve high repeatability score for this negligible image transformation. As it turns out, MSER only manages a repeatability value of only 28.17% for the image pair shown, which is much less than what is expected of the feature detection unit and highlights its unreliable behavior – a stark contrast to the power supply design example.

Having gone through the above two examples, the obvious question which springs to mind is: what is the distinguishing factor between the two approaches which makes one designed system highly reliable and the other one unpredictable? The answer is fairly simple. Every component of an electronic system has known operating characteristics (or performance limits). On the other hand, how much do we all know about the upper and lower performance bounds of the MSER detector, essentially the primary stage of the designed vision system, for 20% uniform decrease in illumination? Not more than what is reported in [1], which is based on a single dataset!

Building reliable and effective vision systems is the sole purpose of working in the area and perhaps this needs to be reiterated. Recently, it has been commented by [4] that the vision community places too much emphasis on beating latest benchmark numbers regardless of whether the improvement over other methods is statistically significant. This is especially true for local feature detection where most methods seem to have confined themselves to some particular datasets for demonstrating their superiority over other competing algorithms. This is essentially where the original objective of building a reliable vision system is lost. The authors contend that it is time to shed the *best-case* analysis approach to concentrate on the original purpose. For achieving the goal of reliability and effectiveness, the operating characteristics of every component in the vision system should be well known. Clearly, a principled framework, utilizing suitable metrics that mirror real-world performance of different components of the system, is required for this – something which is currently lacking.

This paper attempts to bridge this research gap. Limiting itself to the feature detection stage, the paper demonstrates how the improved repeatability measure, proposed in [5], which provides results that are reliable and consistent with the actual performance of a wide variety of detectors across a number of well-established datasets [5], can be utilized from a vision systems design perspective. This paper proposes a generic framework for finding the operating and guarantee regions of a local feature detector under some specific image transformation. Taking into account the comments of [4], the framework also identifies statistically significant performance differences between detectors by introducing a variant of McNemar's test [6, 7]. To demonstrate the utility of this framework, three specific image transformations, namely JPEG compression, blurring and uniform light changes, are used. The operating and guarantee regions for several state-of-the art detectors are established and statistical performance differences are identified for these image transformations. These detailed results are obtained utilizing a newly acquired, large image database (20482 images with 539 different scenes). The results provide insights into the behavior of detectors, and are also useful from the vision systems design perspective.

The paper is organized as follows. Section 2 provides an overview of the related work in the domain of local feature detection evaluation. Section 3 proposes the generic framework for finding the operating and guarantee regions of a given local feature detector under a specific transformation, and for identifying statistical performance differences. The newly acquired, large database of images is introduced in Section 4. The results are presented and discussed in Section 5 and Section 6. Finally, the conclusions are given in Section 7.

## 2. Related Work

This section provides a review of the performance measures presented and evaluation done so far regarding local feature detectors. Corner detectors are evaluated based on chain coded curves by [8]. Visual inspection is used for performance characterization of detectors in [9]. An evaluation of feature detectors based on a quantitative measure of the quality of detected dominant points is performed in [10]. In [11], the localization accuracy of interest point detectors is utilized as a performance measure for comparing them using different planar projective invariants for which reference values are computed using scene measurements. Three interest point detectors are evaluated by utilizing a L-corner model in [12]. In the same spirit, theoretical analysis of L-corners with aperture angles in the range 0–180° is used for comparing feature detectors in [13]. Alignment of the extracted points, accuracy of the 3D reconstruction, accuracy of the epipolar geometry and stability of the cross-ratio are used as criteria to measure the localization accuracy of a model-based L-corner detector by [14]. An approach similar to [11] is utilized in [15] for comparing feature detectors.

A metric based on visual inspection is employed in [16] for evaluating feature detectors. Canny's criteria [17], namely good detection, good localization and low response multiplicity are used as performance metrics for theoretical evaluation of edge detectors by [18]. In [19], structure from motion is used as a specific task to characterize performance. Edge detectors are compared for object recognition task by [20]. Human marked ground-truth is utilized in [21] for assessing the performance of edge detectors. Collinearity, intersection at a single point, parallelism and localization on an ellipse are used as criteria for performance characterization of detectors in [22]. In [23], an evaluation of the quality of detection is carried out based on a set of visual inspection criteria.

Repeatability and information content are utilized as performance metrics in [24]. These two measures are also used in [25] for evaluating feature detectors in the context of image retrieval. The definition of repeatability was refined by [26] and used for evaluating six state-of-the-art local feature detectors in [1]. Improved repeatability measures are presented in [5]. Consistency of the number of corners and accuracy are employed as performance metrics in [27]. The same approach is used by [28] for performance characterization of corner detectors. For evaluating performance of detectors in [29, 30], the number of frames over which the corners are detected during tracking is used as a measure. An evaluation of local feature detectors on non-planar scenes is carried out in [31, 32]. The performance of local feature detectors is compared based on image coverage in [33, 34].

For the specific task of matching 3D object features across viewpoints and lighting conditions, an assessment of the performance of feature detectors is done by [35, 36]. Feature detectors are compared for recognition task using object category training data in [37, 38]. For a pedestrian detection task, an evaluation of feature detectors is done in [39].

In the context of automatic image orientation systems, the performance characterization of local features is carried out by [40]. Localization accuracy of feature detectors is evaluated in [41]. A similar approach based on localization accuracy is reported in [42]. Completeness of detected features is used as a performance metric in [43, 44] for comparing state-of-the-art local feature detectors. In [45], the performance of detectors is evaluated under viewpoint, scale and light changes by using a large database of images with recall rate as performance measure.

The literature on performance metrics and evaluation of local feature detectors based on them is vast and has grown rapidly after the emergence of SIFT [46, 47]. There have been a number of evaluations based on specific vision tasks such as visual SLAM [48] and face detection [49]. It is not possible to describe every such contribution here but an attempt has been made to mention the main developments which are considered important in this domain.

## 3. The Proposed Framework

This section presents a generic framework targeting the feature detection stage of the vision system design process for developing systems that would follow the design specifications and would be more reliable and effective. Before discussing the details, it is worth stating that the proposed framework is based on two main principles:

a) The ability to determine the upper and lower performance bounds of a given detector under some specific type and amount of image transformation — an idea borrowed from electronic systems design practice.

b) The ability to identify statistically significant performance differences between a given detector and some other detector whose performance is considered a benchmark under specific type and amount of image transformation — a concept for taking into account the comment made by [4].

To adhere to the above-mentioned principles, the framework is divided into two distinct components. The details of these components are given below.

### 3.1. Component 1

The first part of the framework establishes the upper and lower performance bounds of a given detector. For achieving this objective, it utilizes the improved repeatability measure presented in [5] which provides results that are reliable and consistent with the actual performance of a wide variety of detectors across a number of well-established datasets:

$$Improved\ Repeatability = \frac{N_{rep}}{N_{ref}} \tag{1}$$

where $N_{rep}$ is the total number of repeated points and $N_{ref}$ is the total number of interest points in the common part of the reference image.

Assuming the availability of a large image database involving a specific type of image transformation with known ground truth mapping between images and consisting of $n$ individual datasets with each having a different scene, the first component of the framework carries out the following steps:

a) The repeatability scores are computed using Equation (1) for all images in every individual dataset (of the large image database) by taking the first image in each dataset which contains no transformation, as the reference. Assuming that the amount of image transformation is varied in $m$ discrete steps for every single dataset, $n$ values of repeatability are obtained for each discrete step. Let $A$ be the set of $m$ discrete steps representing specific transformation amounts

$$A = \{1, 2, 3, \dots \dots, m\} \tag{2}$$

Let $B_k$ be the set of $n$ repeatability values at any one specific step $k$, where $k$ is an element of set $A$

$$B_k = \{b_{1k}, b_{2k}, b_{3k}, \dots \dots, b_{nk}\} \tag{3}$$

For example, if the image database consists of 539 different datasets (the number which will be used in the next few sections), each consisting of a sequence of 14 images, the values of $n$ and $m$ will be 539 and 14 respectively. In other words, there will be 539 values of repeatability available for each step of image transformation amount.

b) For every discrete step $k$, the maximum value of repeatability is

$$P = \{\max(B_1), \max(B_2), \max(B_3), \dots\dots, \max(B_m)\} \tag{4}$$

The values of set $P$ are plotted against the corresponding image transformation amounts from set $A$ to obtain a curve which represents the upper bound of performance for the given detector with variation in the amount of transformation. This curve is named the *max curve*.

c) For every discrete step $k$, the minimum value of repeatability is found to give

$$Q = \{\min(B_1), \min(B_2), \min(B_3), \dots\dots, \min(B_m)\} \tag{5}$$

The values of set $Q$ are plotted against the corresponding image transformation amounts from set $A$ to obtain a curve which represents the lower bound of performance for the given detector with the same variations in image transformation. This curve is named the *min curve*.

d) For every discrete step $k$, the median value of repeatability is found

$$S = \{\mathrm{median}(B_1), \mathrm{median}(B_2), \dots., \mathrm{median}(B_m)\} \tag{6}$$

The values of set $S$ are plotted against the corresponding image transformation amounts from set $A$ to obtain a curve which represents the typical performance for the given detector with variation in image transformation amount. This curve is named the *median curve*.

e) By plotting the three curves together, the area between the *max curve* and the *min curve* is defined as the *operating region* of the detector. The detector is expected to produce repeatability scores that lie inside this region. A narrow o*perating region* implies that the detector is stable and there is little variation between the maximum and minimum repeatability values that it can achieve for some specific amount of transformation. On the other hand, a large *operating region* indicates an unstable detector which may achieve high repeatability scores for some particular images but may fare poorly for others.

f) The area under the *min curve* is defined as the *guarantee region* of the detector. Repeatability values achieved by the detector should never be as low so as to lie inside this region. A wide *guarantee region* shows that the detector manages to achieve reasonably high repeatability values for every input image with increasing amount of image transformation. Contrary to that, a small *guarantee region* implies that the detector performs poorly with increasing amount of image transformation.

### *3.2. Component 2*

The second part of the proposed framework identifies statistically significant performance differences between two given detectors by introducing a variant of the non-parametric McNemar's test [6, 7].

McNemar's test is a form of chi-squared test with one degree of freedom that evaluates the performance of two algorithms based on their outcomes on a case-by-case basis over the same dataset [6, 7]:

$$Z = \frac{\left|N_{sf} - N_{fs}\right| - 1}{\sqrt{N_{sf} + N_{fs}}} \qquad (7)$$

where $N_{sf}$ and $N_{fs}$ are the numbers of occurrences when one algorithm succeeds and the other algorithm fails. If $N_{sf} + N_{fs} \geq 30$, the statistic is reliable and $Z$ can be converted into a probability using tables [6, 7].

For utilizing McNemar's test, a criterion is needed to determine whether a test case results in success or failure. This framework utilizes repeatability score as this criterion. By selecting a specific threshold value of the repeatability score, it is possible to determine the success or failure of a detector. However, there is a variety of feature detectors available which show large variations in absolute and relative performances for different types of image transformations, so it is difficult to select one specific repeatability threshold that would work for all cases without introducing any bias. To solve this problem, this framework introduces a variant of McNemar's test. Instead of fixing the threshold value for repeatability, this variant utilizes a ROC-like approach, where the value of the threshold is varied in $t$ discrete steps for each specific image transformation amount. Let $X$ be the set of discrete steps that represent specific test thresholds

$$X = \{1, 2, 3, \ldots, t\} \qquad (8)$$

For any value $k$ of set $A$, which represents the image transformation amount, there will be $t$ $Z$-scores obtained for the two given detectors. This may be represented mathematically as:

$$Y_k = \{y_{1k}, y_{2k}, y_{3k}, \ldots, y_{tk}\} \qquad (9)$$

By varying the value of $k$ in the above equation within the range of set $A$ allows $Z$-scores obtained for the two detectors to be viewed as a function of image transformation amount and test threshold. This can conveniently be displayed in the form of an image.

## 4. The Image Database

This section presents a newly acquired image database for finding the performance bounds of different local feature detectors. With 539 different scenes, the database contains 20482 images involving various image transformations, namely JPEG compression, blurring and uniform light changes. Some images from the image database are shown in Figure 2. Each image in the database consists of 717 x 1080 pixels. To facilitate future research in this area, the image database is made available at [50].

### 4.1. JPEG Compression

In [1], the authors examined the performance of different local feature detectors on the basis of a single dataset, UBC [2], for JPEG compression ratios varying from 60% to 98%. To investigate the behavior of local feature detectors by employing the framework proposed in the previous section reliably, a much larger database of images with variation in JPEG compression ratio is required. Since there is no such resource available, this section presents a newly acquired database of images with JPEG compression ratios varying from 0% to 98%. The database consists of 7546 images with 539 different planar scenes, captured by the authors; both structured and textured scenes are included to eliminate any potential dataset bias. It should be noted that the maximum number of scenes which has been used so far for studying the performance of local feature detectors under different image transformations is only 60 [45]; the number of scenes for the presented database is thus nearly 9 times that of what is used in [45]. Moreover, the scenes employed in [45] are not real-world scenes and are captured in a highly controlled environment, whereas the presented database consists of images with scenes that are encountered routinely in everyday life (see Figure 2).

For every scene, the JPEG compression ratio is varied in 14 discrete steps from 0% to 98% (14 x 539 = 7546). The database was generated using the cjpeg and djpeg utilities in a Linux-based environment by varying the image quality parameter. Since there is no geometric transformation involved in the case of JPEG compression, the ground truth homography relating any two images of the same scene with different compression ratios is simply a 3 x 3 identity matrix.

*4.2. Blur Changes*

For investigating the behavior of local feature detectors utilizing the proposed framework under blur changes, a large database of images involving the same scenes as in the JPEG compression case is presented with variations in the amount of blur. It should be noted that only two datasets of six images each, one containing a textured and the other a structured scene (Trees and Bikes datasets respectively [2]), are used in [1] to determine the performance of six state-of-the-art local feature detectors. Instead, the presented database consists of 5390 images in total with 539 different planar scenes captured by the authors. Both structured and textured real-world scenes are included to ensure that there is no bias shown towards any particular detector when determining the upper and lower performance bounds.

The amount of blur is varied in 10 discrete steps for each scene (10 x 539 = 5390). The database has been generated digitally utilizing MATLAB; the first image of every scene (having no blur) is convolved repeatedly with Gaussian blur kernels, having the same size as the image, with increasing standard deviations to produce a sequence of images with increasing amount of blur. More specifically, standard deviations ranging from 0.5 to 4.5 with a step size of 0.5 are used for the blur kernels. Since the increasing amount of blur does not cause any geometric transformation in the image with respect to the previous images in the same sequence, the ground truth homography which provides the image-to-image mapping for any two images of the same scene with different amounts of blur is simply a 3 x 3 identity matrix.

*4.3. Uniform Light Changes*

Among the Oxford datasets [2], only Leuven, consisting of a sequence of six images, involves uniform changes in light. In [45], a large image database is presented to investigate the effect of light direction on the performance of feature detectors. However, the total number of scenes that have been used in that database is only 60. Also, the scenes employed are not real-world scenes but are captured in a highly controlled environment. A large image database is thus presented here to investigate the behavior of local feature detectors under uniform changes in illumination by employing the proposed framework. The database consists of 7546 images, involving the same 539 scenes as in the JPEG and blur image databases, with variation in illumination. Both structured and textured real-world scenes are included. The number of scenes for the presented database is nearly 9 times that of what is used in [45].

For every scene, the brightness level is decreased in 14 discrete steps from 0% to 90% (14 x 539 = 7546). The database has been generated digitally using MATLAB by varying the image brightness level. The ground truth homography that relates any two images of the same scene with different light conditions in the presented database is a 3 x 3 identity matrix as uniform changes in light do not result in any geometric transformation.

**Figure 2.** Some sample scenes from the image database.

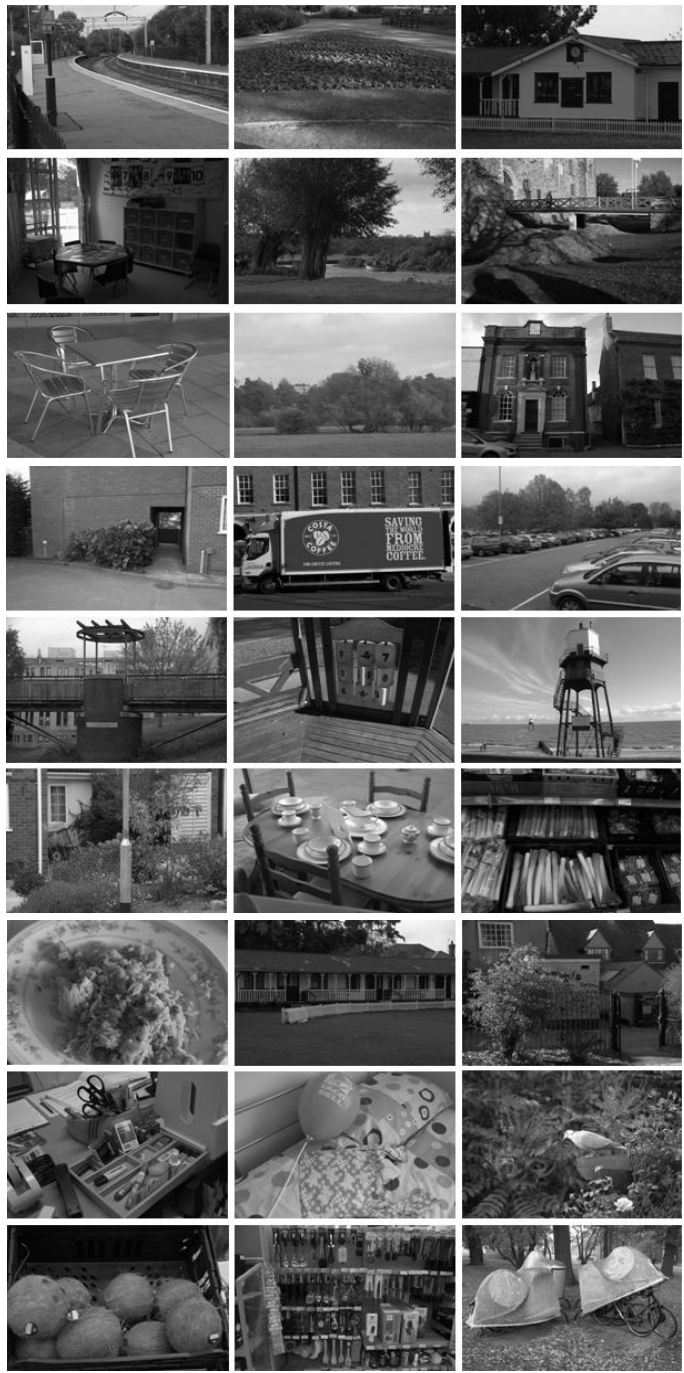

## 5. Establishing Operating and Guarantee Regions

This section establishes *operating* and *guarantee* regions for several state-of-the-art feature detectors for JPEG compression, blurring and uniform light changes utilizing the proposed framework and the large image database. These detectors include SIFT [46, 47], SURF [51], Harris-Laplace, Hessian-Laplace, Harris-Affine, Hessian-Affine [26], Intensity-based Regions (IBR), Edge-based Regions (EBR) [52], Maximally Stable Extremal Regions (MSER) [3], SFOP [53] and Salient [54]. These were chosen because they are scale- and rotation-invariant detectors and representative of a number of different approaches to feature detection [55]; also their implementations are widely available. Although the control parameters of these feature detectors can be varied to yield a similar number of interest points for all detectors, this approach has a negative effect on their repeatability and performance [26]. Therefore, authors' original programs (binary or source) have been utilized with parameters set to values recommended by them. The parameter settings used make these results a direct complement to existing evaluations.

### 5.1. JPEG Compression

Results for several state-of-the-art detectors utilizing the framework proposed in Section 3 are presented in Figure 3 and Figure 4. These results determine the upper and lower bounds of performance of detectors with varying JPEG compression ratio, and then establish their operating and guarantee regions. This approach of presenting performance limits is intended to provide information for the design of robust vision systems; it is entirely possible that the standard deviations of performance may be significantly smaller than these limits. Before discussing the results, it is worth stating that this appears to be the first attempt to do such a detailed analysis; there is no other work in the literature with which these findings can be compared to determine consistencies and contradictions. The results provide useful insight into the behavior of detectors under JPEG compression.

Figure 3 and Figure 4 depict the operating and guarantee regions for several state-of-the-art detectors. Although MSER does perform well for some particular images with increasing JPEG compression ratio (see the max curve of MSER in Figure 3), it is evident from the large operating region that its behavior is unstable. Even for small amounts of transformation, MSER fails to achieve high repeatability values for some images (see the value of min curve at 20% JPEG compression). Such an unpredictable performance does not make MSER a suitable choice for vision systems with more than 10% JPEG compression ratios as the detector may perform poorly for some images encountered.

IBR is more stable than MSER with increasing JPEG compression ratios as its operating region is smaller. It is however noticeable that IBR fails to beat the max curve of MSER. Although reasonable values of repeatability are achieved by IBR for JPEG compression ratios up to 95%, the performance may go to nearly zero for some particular images for 98% compression ratio.

SURF performs well for increasing JPEG compression ratios up to 95% as is evident from its wide guarantee region. It shows relatively poor stability only for the case when JPEG compression ratio is 98%. This makes SURF a good choice from a vision systems design perspective when the expected JPEG compression ration does not exceed 95%.

**Figure 3.** JPEG compression results utilizing the proposed framework and the image database for MSER (top left), IBR (top right), Harris-Laplace (center left), Hessian-Laplace (center right), SURF (bottom left) and SIFT (bottom right) detectors.

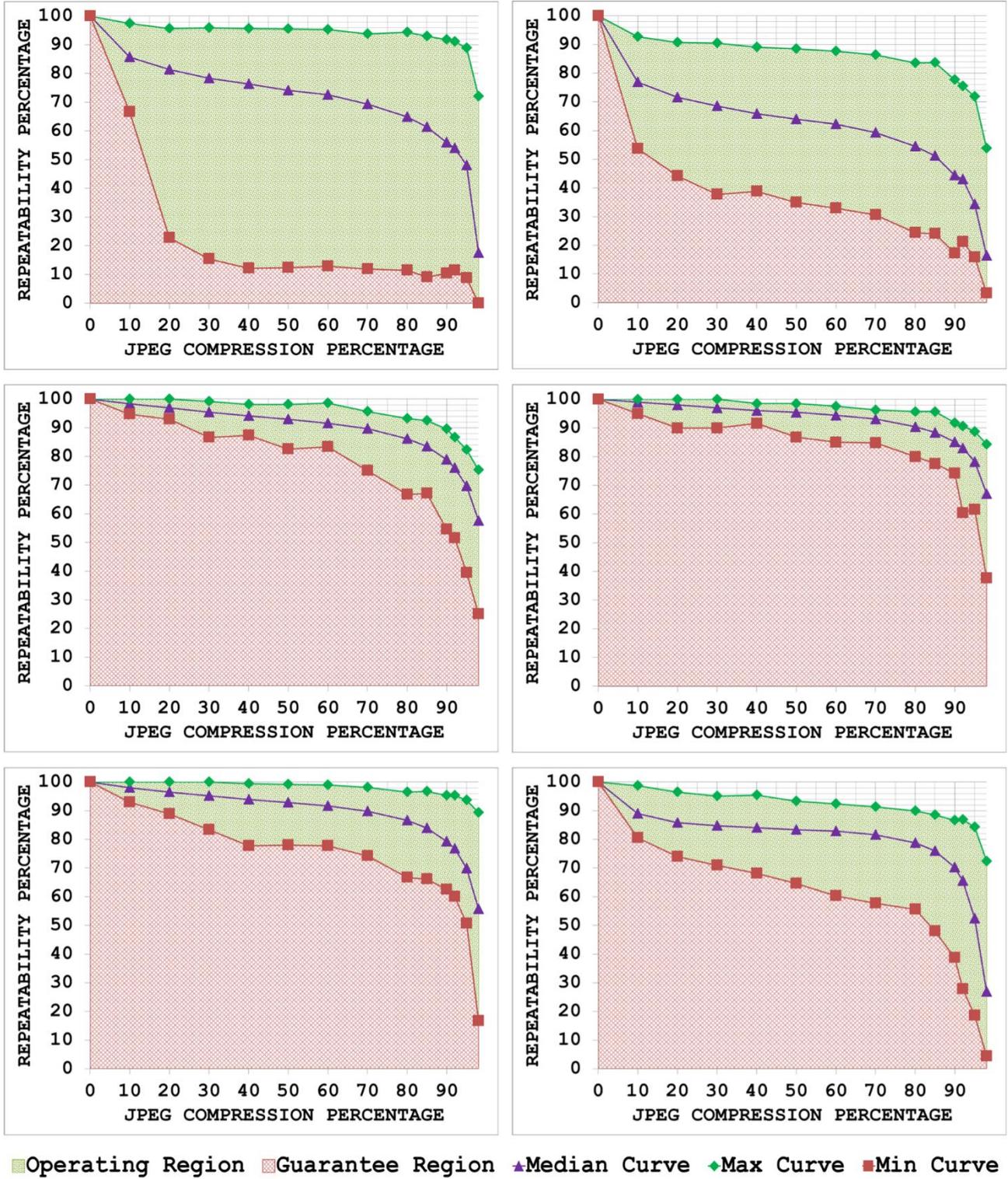

The results for Harris-Laplace and Hessian-Laplace utilizing the proposed framework are also shown in Figure 3. The narrow operating regions for these detectors, especially Hessian-Laplace, demonstrate their stability to increasing JPEG compression. These two detectors also have wide guarantee regions, which indicate that they manage to achieve high repeatability scores even for large JPEG compression ratios. Although the performance of these detectors fall sharply when the compression ratio exceeds 85%, the repeatability scores are still reasonable compared to all other detectors but SURF.

Although the performance of SIFT detector is reasonable, its operating region is wider than those of Hessian-Laplace and grows with increasing JPEG compression ratio. Moreover, the performance of SIFT may go to nearly zero depending upon the image content for 98% compression ratio (see the min curve of SIFT in Figure 3). It may be concluded that SIFT is a suitable option for vision systems expecting JPEG compression ratios up to 50%.

Figure 4 shows the results for Salient utilizing the proposed generic framework. A much wider guarantee region for JPEG compression ratios up to 20% indicates that Salient is relatively more stable compared to MSER and IBR up to this point. For ratios greater than that, the operating region for Salient becomes wider and the min curve nearly goes to zero for 98% JPEG compression ratio. From a vision systems design perspective, Salient is a not an appropriate option if the JPEG compression ratio is expected to be more than 20%.

It is evident from Figure 4 that EBR shows highly unstable behavior. The operating region is wide, indicating that the performance of EBR may vary between rather high and rather low repeatability values for increasing JPEG compression ratios, depending upon the image content. Again, such behavior is not desirable when designing vision systems as it jeopardizes the final output of the system. Thus, EBR does not appear to be the best detector, even for vision systems expecting small JPEG compression ratios.

Figure 4 also depicts the results for SFOP. It is clear that the detector experiences small but continuous degradation in performance with increasing JPEG compression ratios; indeed, its narrow operating region for compression ratios up to 60% indicates that the detector is quite stable up to this point. For JPEG compression ratios exceeding 90%, SFOP may fail to achieve repeatability scores greater than 35%, depending upon the image content. Finally, the results for Harris-Affine and Hessian-Affine utilizing the proposed framework are presented in Figure 4 which are similar to their counterparts (Harris-Laplace and Hessian-Laplace in Figure 3).

Not only do the results shown in Figure 3 and Figure 4 for JPEG compression provide insights into the behavior of detectors for this particular transformation, they also allow a vision system designer to select the best detector considering the design constraints for achieving more reliability — a feature inherited from electronic systems design practices.

**Figure 4.** JPEG compression results utilizing the proposed framework and the image database for Salient (top left), EBR (top right), Harris-Affine (center left), Hessian-Affine (center right), SFOP (bottom) detectors.

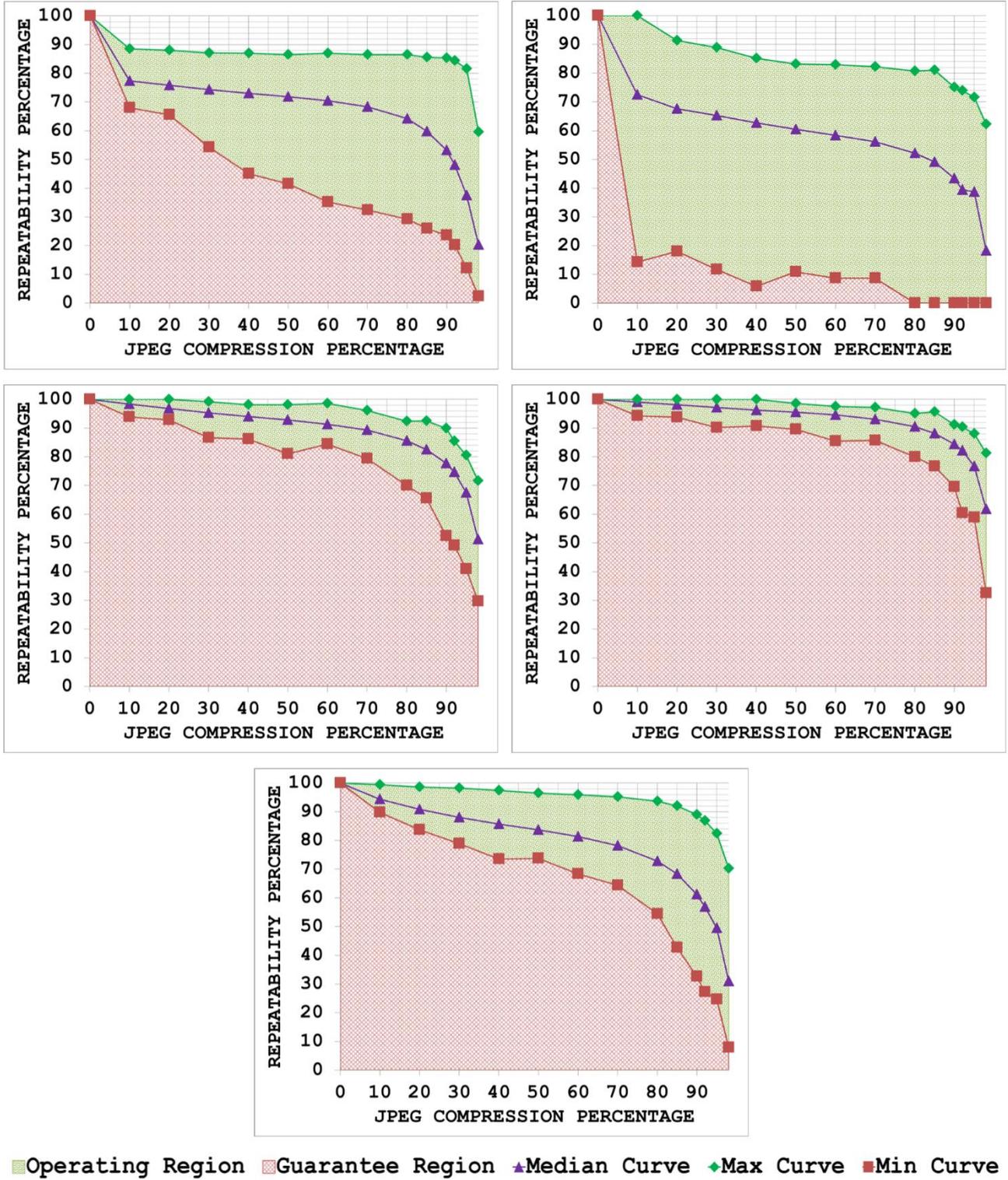

## 5.2. Blur Changes

Figure 5 and Figure 6 present results for several state-of-the-art feature detectors utilizing the proposed generic framework which establish their operating and guarantee regions under changes in blur. Again, there is no other detailed work in the literature with which these results can be compared to determine the consistencies and contradictions. In [1], the authors had concluded on the basis of only two datasets that the detectors under examination are robust to changes in blur, as they featured almost horizontal repeatability curves. The results presented here are more comprehensive and largely contradict that perception. It should be noted that SIFT detects more than 20,000 features for some images in the blur image database which makes it very time-consuming to do such a detailed analysis for SIFT. In the case of JPEG image database, it took more than two months to obtain results on HP ProLiant DL380 G7 system with Intel Xeon 5600 series processors. Therefore, results for SIFT are not provided in this section.

The results for MSER and IBR are shown in Figure 5. It is evident that the two detectors undergo a decline in performance with increasing amount of blur. Both MSER and IBR are segmentation-based detectors but it is interesting to note that the guarantee region of IBR is much wider than that of MSER. Moreover, the operating region of MSER is large, indicating that the detector is unstable; it may provide high repeatability scores for some particular images yet may fare poorly for others. Such unpredictable behavior is not suitable from the vision systems design viewpoint.

The operating and guarantee regions for SURF is also shown in Figure 5. It is clear that the operating region of SURF grows rapidly with increasing blur, thus indicating unpredictable behavior of the detector. Depending upon the image content, SURF may fail to provide repeatable features in the presence of increasing blur (see the min curve in Figure 5). Conversely, SFOP shows comparatively good performance with a large guarantee region and a narrow operating region (see Figure 5). It seems quite stable under increasing blur and the max and min curves are also fairly smooth, indicating a gradual degradation in performance. Figure 5 and Figure 6 depict results for Harris-Laplace, Hessian-Laplace, Harris-Affine and Hessian-Affine. The operating and guarantee regions of these four detectors are similar, although Hessian-based detectors appear better than Harris-based ones. For small amounts of blur, the detectors demonstrate good performance but may fare poorly in the presence of increasing blur.

Figure 6 shows the operating and guarantee regions for Salient. There is continuous degradation in the performance of Salient with increasing blur. However, the detector appears more stable than MSER and IBR, as indicated by a narrow operating region. The results for EBR are also depicted in Figure 6. The performance of EBR depends largely upon the image content as it may achieve good repeatability scores for some image while its performance may go to zero for others. A large operating region for EBR points to its unstable behavior under changes in the amount of blur.

*5.3. Uniform Light Changes*

Figure 7 and Figure 8 depict the upper and lower performance bounds of several state-of-the-art feature detectors under uniform changes in light utilizing the proposed generic framework. Again, the results of SIFT are not provided as it detects a large number of features in some images of the database (in excess of 20,000), making the computation time prohibitively large for such a detailed analysis.

In [1], the authors have concluded that the six detectors under study are highly robust to uniform variations in illumination. As mentioned earlier, this deduction is based on a single dataset (Leuven [2]). The results presented here largely contradict those findings, showing that there is a rapid decline in the performance in the presence of uniform light changes. A similar performance degradation effect is observed in [45] while studying the behavior of feature detectors under changes in light direction.

**Figure 5.** Results for blur changes utilizing the proposed framework and the image database for MSER (top left), IBR (top right), Harris-Laplace (center left), Hessian-Laplace (center right), SURF (bottom left) and SFOP (bottom right) detectors.

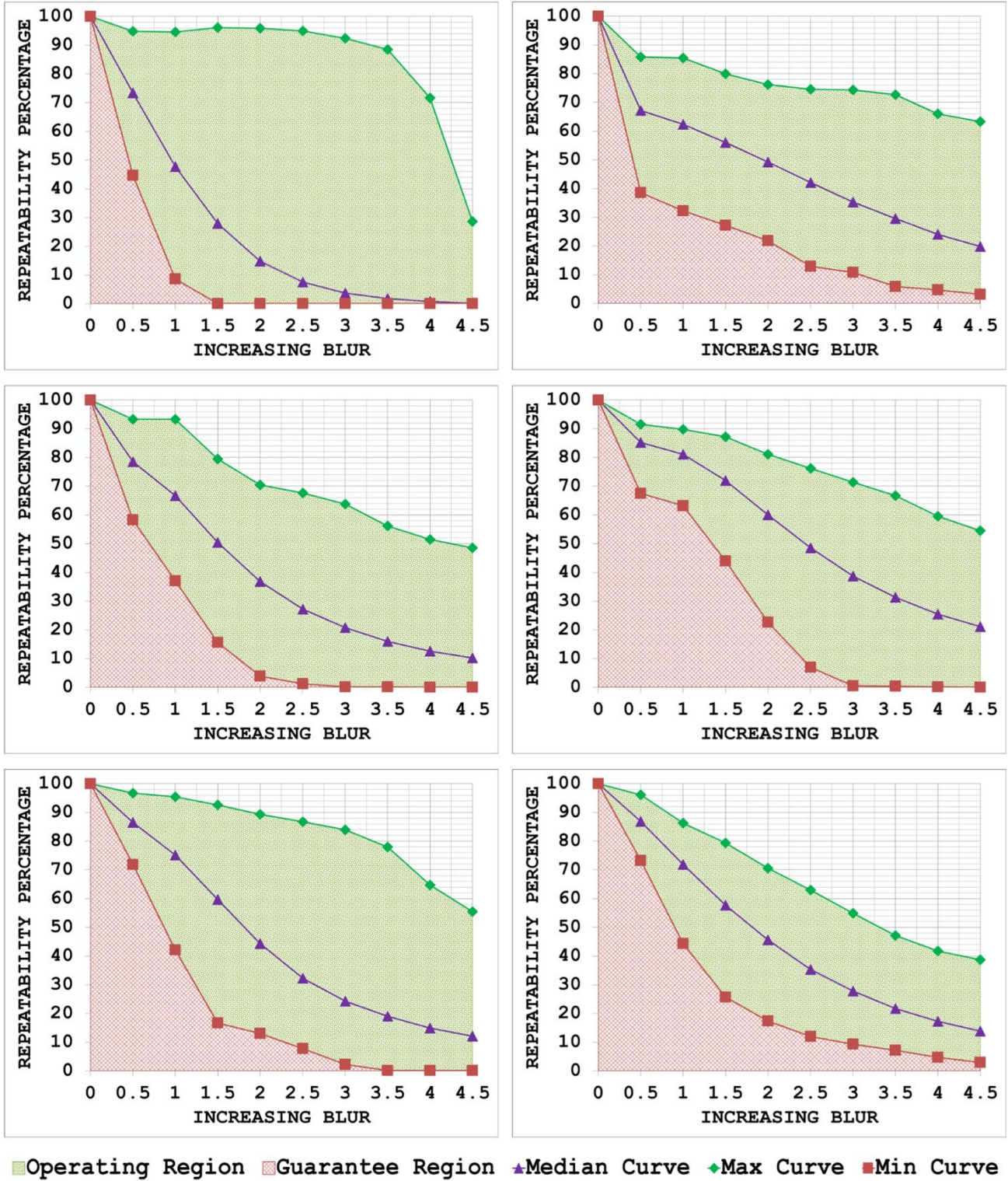

**Figure 6.** Results for blur changes utilizing the proposed framework and the image database for Salient (top left), EBR (top right), Harris-Affine (bottom left) and Hessian-Affine (bottom right) detectors.

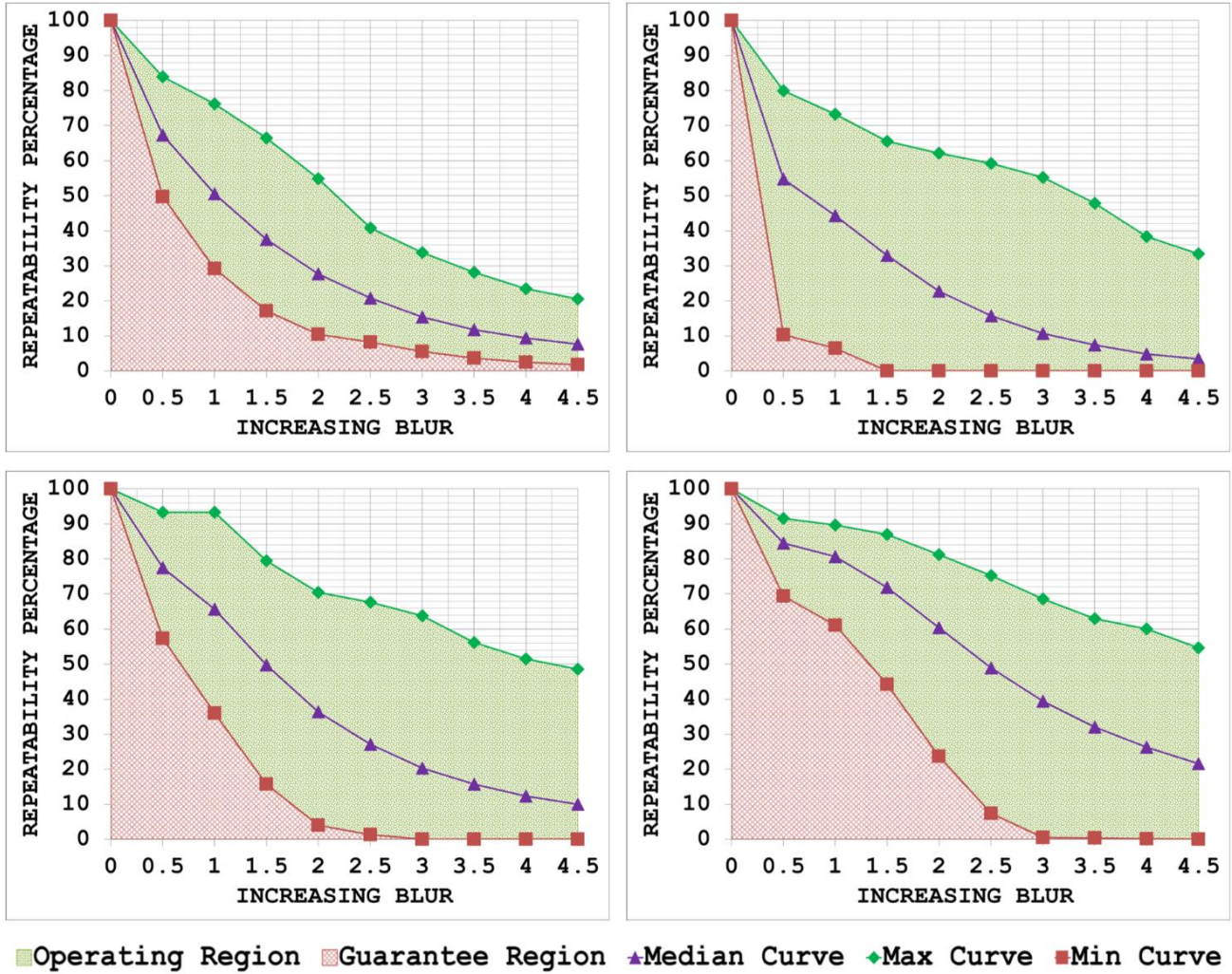

Figure 7 shows the results for the two segmentation-based detectors, MSER and IBR, respectively. It is evident that both the detectors have very large operating regions which indicate their unstable behavior in the presence of decreasing light. It is interesting to note that the min curve of MSER, the detector which is identified as the best for this specific image transformation in [1], reaches zero for only 50% uniform decrease in light (see Figure 7). MSER and IBR do not seem suitable for vision systems expecting more than 10% uniform decrease in light.

There is a rapid decline in the performance of SURF with decreasing light (see Figure 7). SFOP seems to perform much better as indicated by its wide guarantee region. However, its operating region shows that the performance of the detector may vary between high and low values of repeatability. The results for Harris-Laplace, Hessian-Laplace, Harris-Affine and Hessian-Affine are depicted in Figure 7 and Figure 8. It is clear that all these detectors undergo a quick degradation in performance with decreasing light. The min curves for these detectors show that they hardly manage to achieve a repeatability score of 10-15% in the presence of only 20% uniform decrease in light (see Figure 7 and Figure 8). The operating regions of these detectors are large and their guarantee regions are narrow, meaning that they may achieve high repeatability scores for some images but may fare poorly for others. This large variation in performance for the same amount of image transformation is not desirable from a vision systems design perspective.

**Figure 7.** Results for uniform light variations utilizing the proposed framework and the image database for MSER (top left), IBR (top right), Harris-Laplace (center left), Hessian-Laplace (center right), SURF (bottom left) and SFOP (bottom right) detectors.

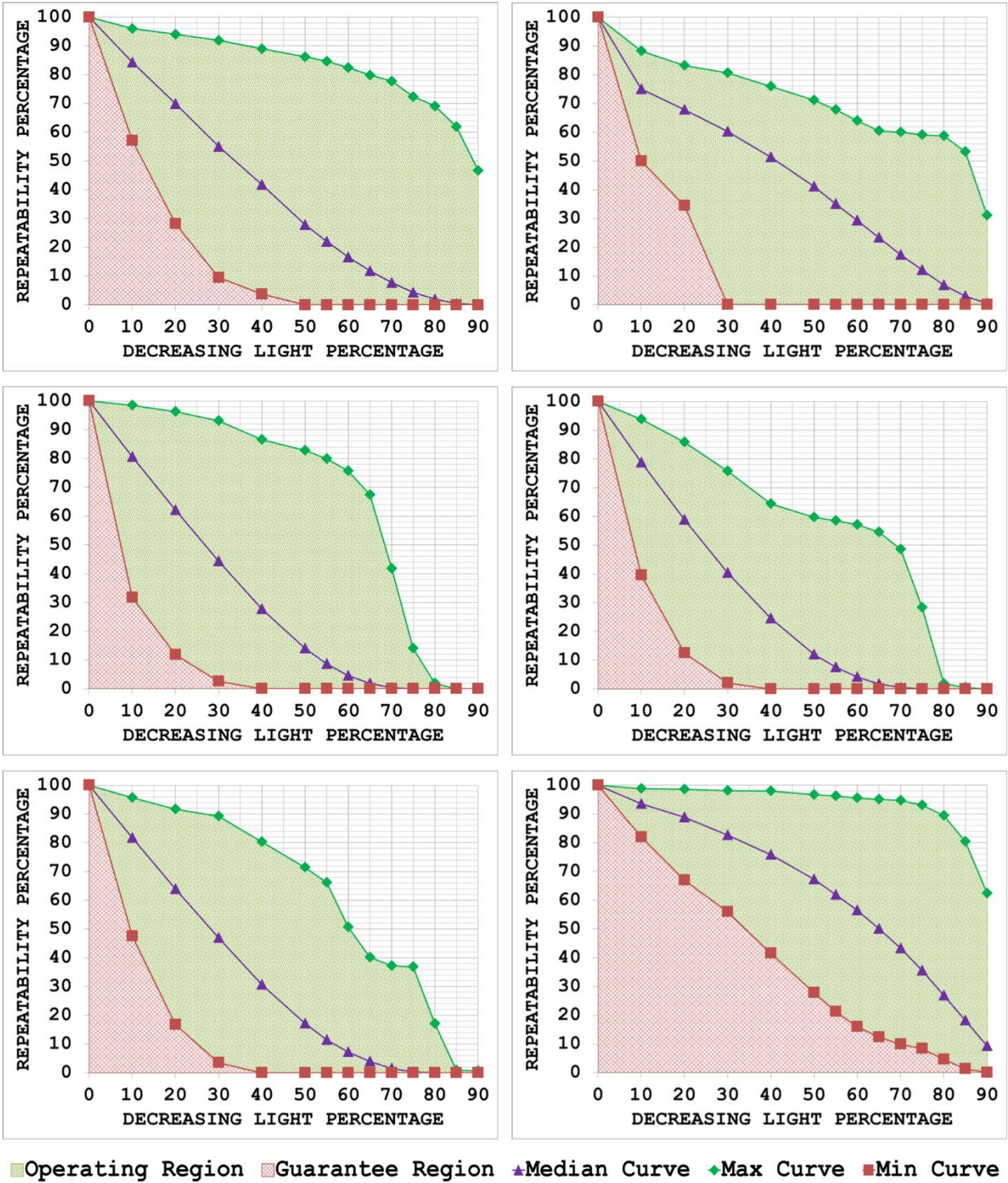

**Figure 8.** Results for uniform light variations utilizing the proposed framework and the image database for Salient (top left), EBR (top right), Harris-Affine (bottom left) and Hessian-Affine (bottom right) detectors.

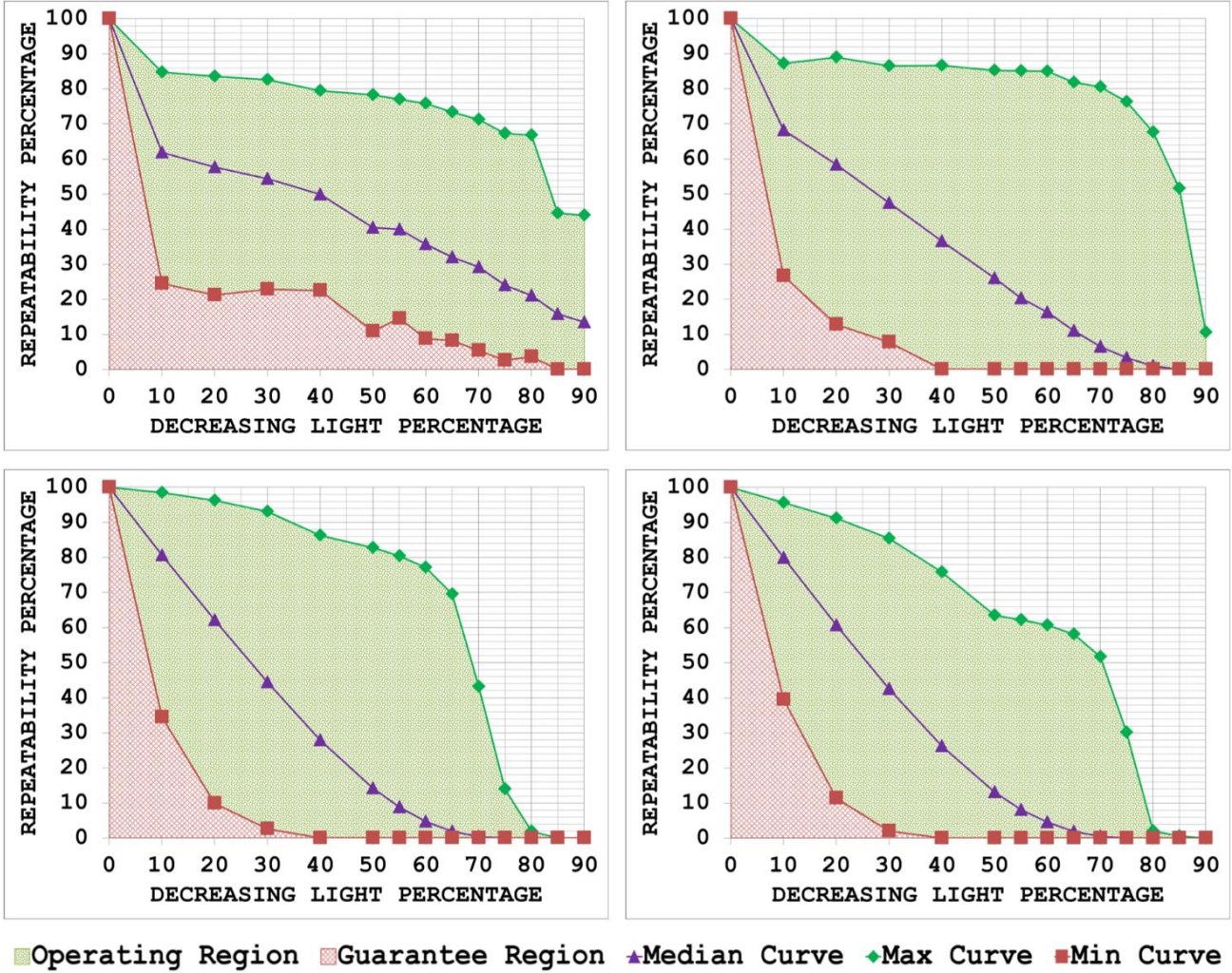

The operating and guarantee regions for Salient and EBR are also depicted in Figure 8. As with MSER and IBR, both Salient and EBR have large operating regions, indicating that they may achieve high repeatability values for some particular images yet may fare poorly for some other images in the presence of uniform changes in illumination; such unpredictable behavior is not desirable from a vision systems design perspective.

## 6. Identifying Statistically Significant Performance Differences

Utilizing the proposed framework and the newly acquired image database, this section carries out the statistical performance comparison of the state-of-the-art feature detectors for various image transformations, namely JPEG compression, blur changes and uniform illumination variations. The results are presented in Figure 9 to Figure 18. Color coding in these figures indicate the *Z-scores* obtained as a function of image transformation amount and McNemar's test threshold when one detector is compared with another. Although the value of $Z$ is always positive, here a sign convention has been used to distinguish the detector with the better performance of the two examined: a positive *Z-score* shows that the first detector is better than the second, whereas a negative value of *Z* indicates the converse.

### 6.1. JPEG Compression

It is evident from Figure 9 that Hessian-Laplace out-performs EBR, IBR, MSER and Salient for most JPEG compression ratios when the test threshold is varied from 10% to 90% as indicated by large positive values of *Z* — a confirmation that the performance differences between Hessian-Laplace and these detectors are statistically significant. A comparison of Hessian-Laplace with Harris-Laplace, Harris-Affine, Hessian-Affine and SURF shows that they have broadly similar performance, although Hessian-Laplace does dominate them for some particular JPEG compression ratios at specific test thresholds. From Figure 9, it can be concluded that Hessian-Laplace also performs better than SFOP and SIFT.

The statistical comparison of SFOP and SIFT is interesting: the two detectors have largely similar performance but for some particular test thresholds and JPEG compression ratios, SFOP out-performs SIFT and vice versa. SURF is dominant when compared to SIFT, Salient and SFOP. It appears that EBR fails to achieve better performance than all other state-of-the-art feature detectors in most cases, as is evident from the large negative values of *Z* in Figure 10. IBR is also comprehensively outperformed by SURF, and to some extent by SFOP and SIFT.

Harris-Laplace and Harris-Affine show better performance than IBR, MSER, Salient, SIFT and SFOP for most test thresholds and JPEG compression ratios in Figure 11. Moreover, their performances are largely similar to each other and to those of SURF and Hessian-Affine. In Figure 12, Hessian-Affine out-performs MSER, IBR, Salient and SFOP. The large positive values of Z for some particular test thresholds and JPEG compression ratios indicate that the performance differences between Hessian-Laplace and SIFT are statistically significant, with the former appearing better of the two compared. SURF, SFOP and SIFT also seem to out-perform MSER. Finally, for some particular test thresholds and JPEG compression ratios in Figure 12, Salient performs better than MSER and vice versa.

**Figure 9.** JPEG compression results for Hessian-Laplace, SFOP and SURF with other detectors showing Z-scores obtained; positive Z-scores indicate that the first detector is better than the second whereas the negative values show the converse.

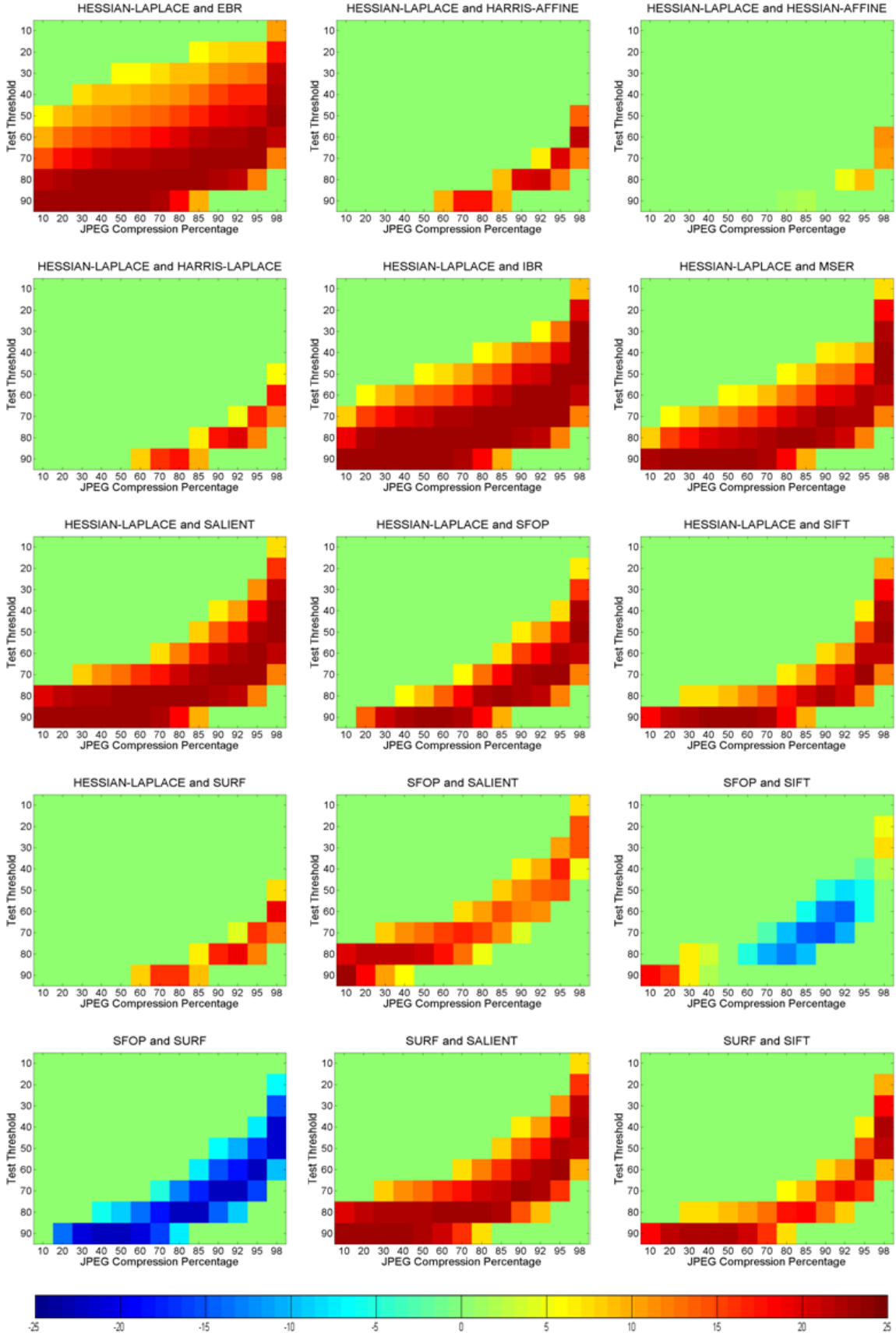

**Figure 10.** JPEG compression results for EBR, IBR and Salient with the other detectors showing Z-scores obtained; positive Z-scores indicate that the first detector is better than the second whereas the negative values show the converse.

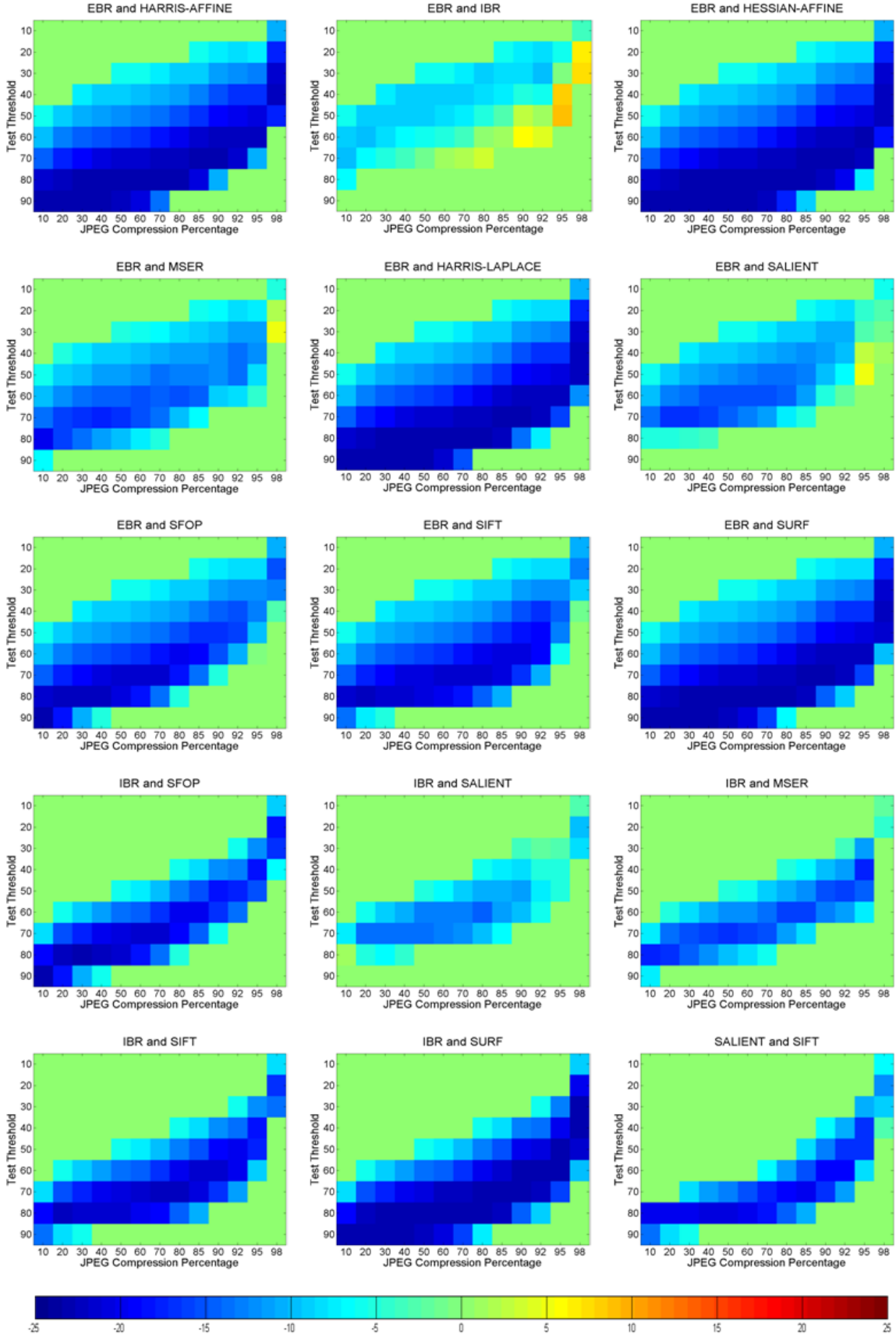

**Figure 11.** JPEG compression results for Harris-Laplace and Harris-Affine with other detectors showing Z-scores; positive Z-scores indicate that the first detector is better than the second whereas negative values show the converse.

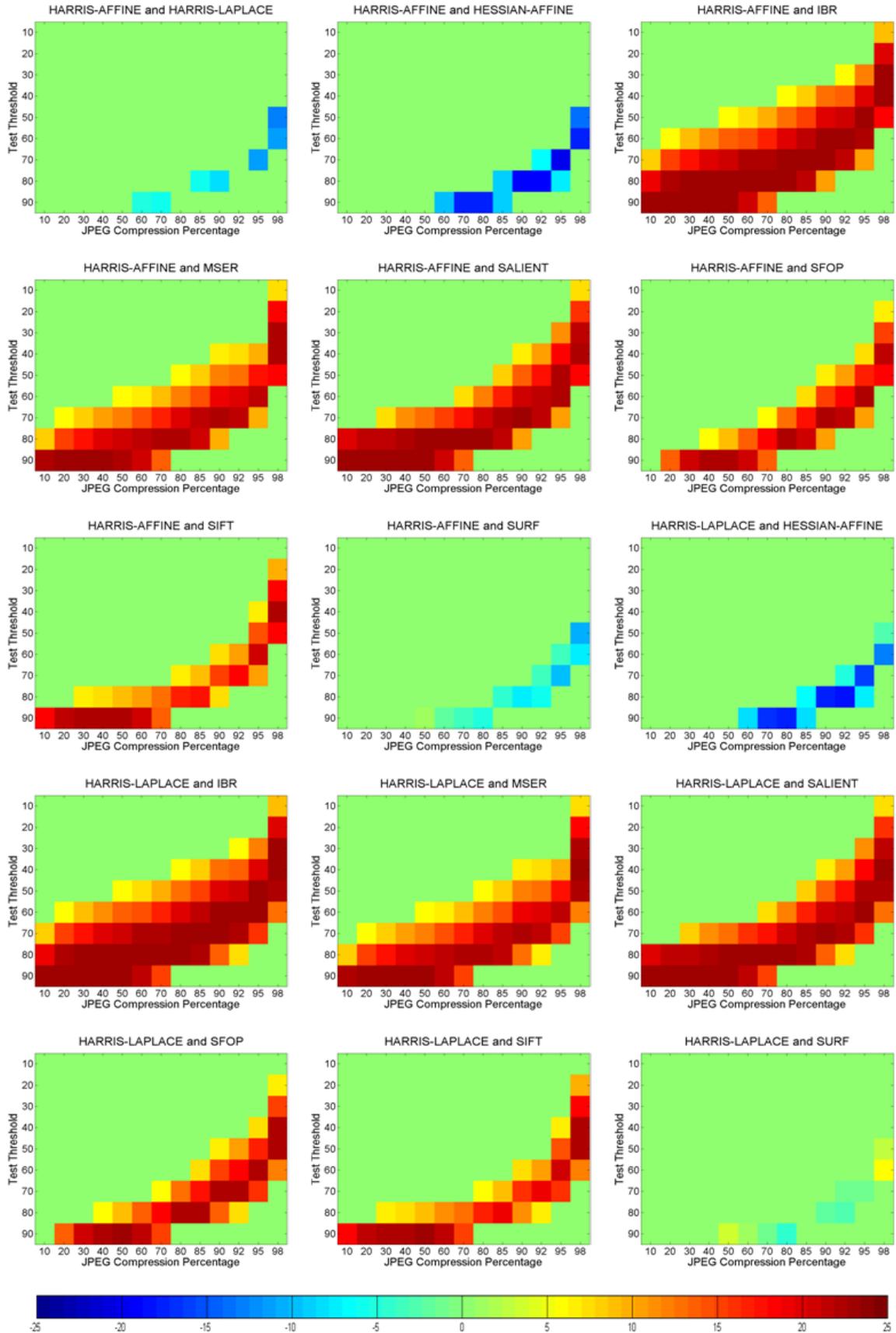

**Figure 12.** JPEG database results for Hessian-Affine and MSER with the other detectors showing Z-scores obtained; positive Z-scores indicate that the first detector is better than the second whereas the negative values show the converse.

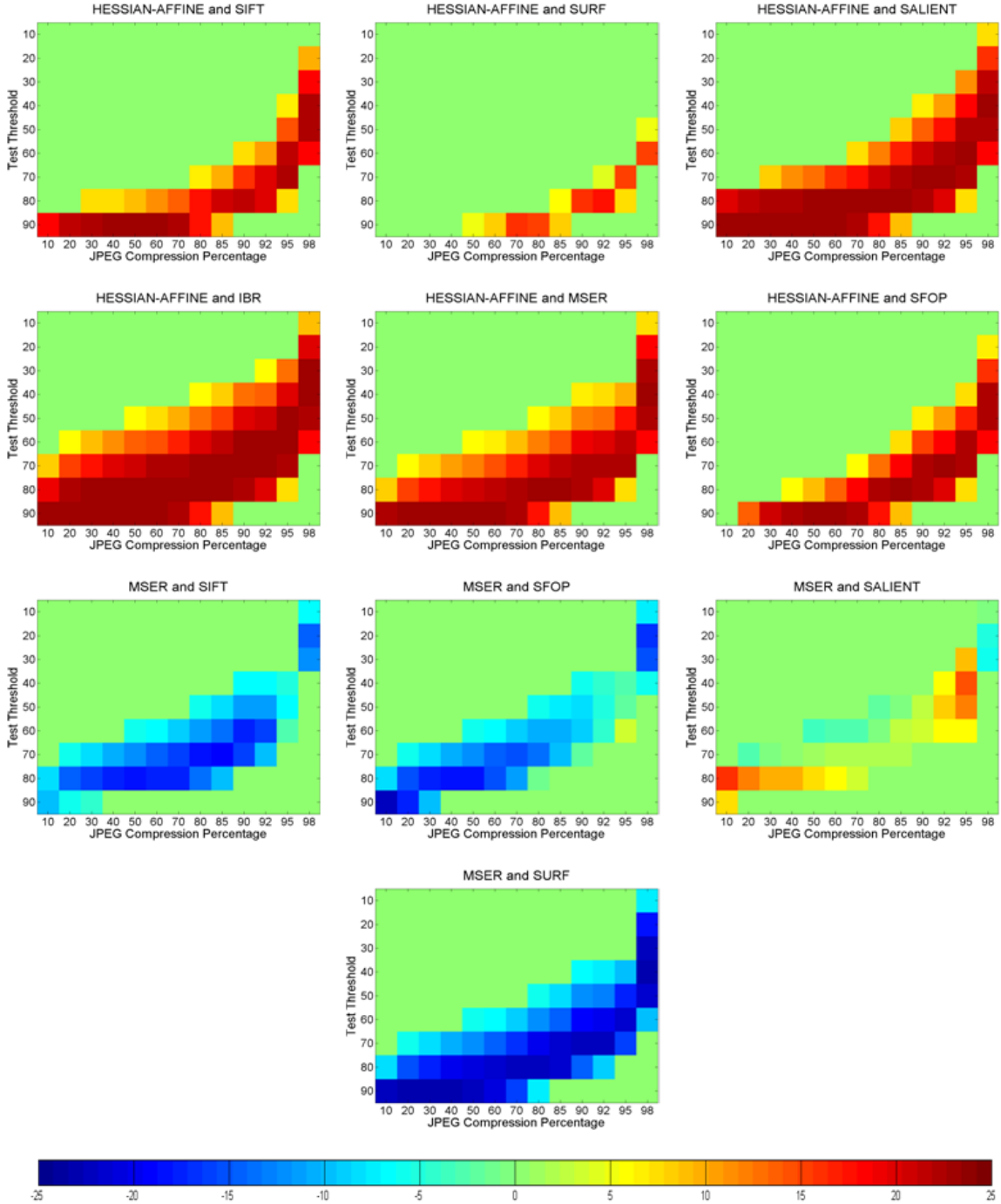

## 6.2. Blur Changes

Figure 13 to Figure 15 depict the results for statistical performance comparison of state-of-the-art feature detectors utilizing the proposed framework under changes in blur. It is clear from Figure 13 that the performance differences between Hessian-Laplace and other detectors, except Hessian-Affine, are statistically significant for most test thresholds and blur amounts, with Hessian-Laplace turning out to be the better detector of the two compared. Harris-Laplace performs better than MSER and Salient (see

Figure 13) but is dominated by Hessian-Affine, IBR, SFOP and SURF for most test thresholds and blur amounts. In Figure 14, it is evident that EBR is comprehensively out-performed by all other detectors but MSER. Harris-Affine and Harris-Laplace show largely similar performances (see Figure 14). The performance of Harris-Affine is better than MSER and Salient in Figure 14, while large negative Z-scores show the supremacy of Hessian-Affine, IBR, SFOP and SURF over Harris-Affine. Like Hessian-Laplace, Hessian-Affine out-performs IBR, MSER, Salient, SFOP and SURF in Figure 15. The performance differences between IBR and MSER are statistically significant, with IBR appearing the better of the two. IBR also performs better than Salient. When IBR is compared with SURF and SFOP, both positive and negative Z-scores are obtained for different test thresholds and blur amounts (see Figure 15). Also, MSER is dominated by Salient, SURF and SFOP, whereas Salient is out-performed by SFOP and SURF.

### 6.3. Uniform Light Changes

Figure 16 shows the performance differences of Harris-Laplace and Hessian-Laplace with the other detectors. It appears that nearly all detectors, including EBR, perform better than Hessian-Laplace for most test thresholds and decreasing light percentages. Apart from Hessian-Laplace and Hessian-Affine, Harris-Laplace fails to out-perform other detectors and is particularly dominated by SFOP, Salient and IBR. EBR shows better performance when compared with Hessian-Affine, Harris-Laplace, Harris-Affine and SURF (see Figure 17). SFOP, Salient and IBR show supremacy over EBR, whereas MSER has largely similar performance to EBR for most test thresholds and decreasing light percentages. As with Harris-Laplace, Harris-Affine is also out-performed by SFOP, Salient, IBR and MSER in Figure 17.

While Hessian-Affine and SURF seem to have largely similar performances, Hessian-Affine fares poorly when compared with IBR, MSER, Salient and SFOP (see Figure 18). Of the two segmentation-based detectors, IBR seems the better. MSER and IBR out-perform SURF but are dominated by SFOP and Salient. The performance differences of SURF with SFOP, and Salient are also statistically significant for most test thresholds and decreasing light percentages, with SURF emerging as the worst of the detectors compared. SFOP also out-performs Salient comprehensively.

**Figure 13.** Results of blur changes for Harris-Laplace and Hessian-Laplace with other detectors showing Z-scores obtained; positive Z-scores indicate that the first detector is better than the second whereas negative values show the converse.

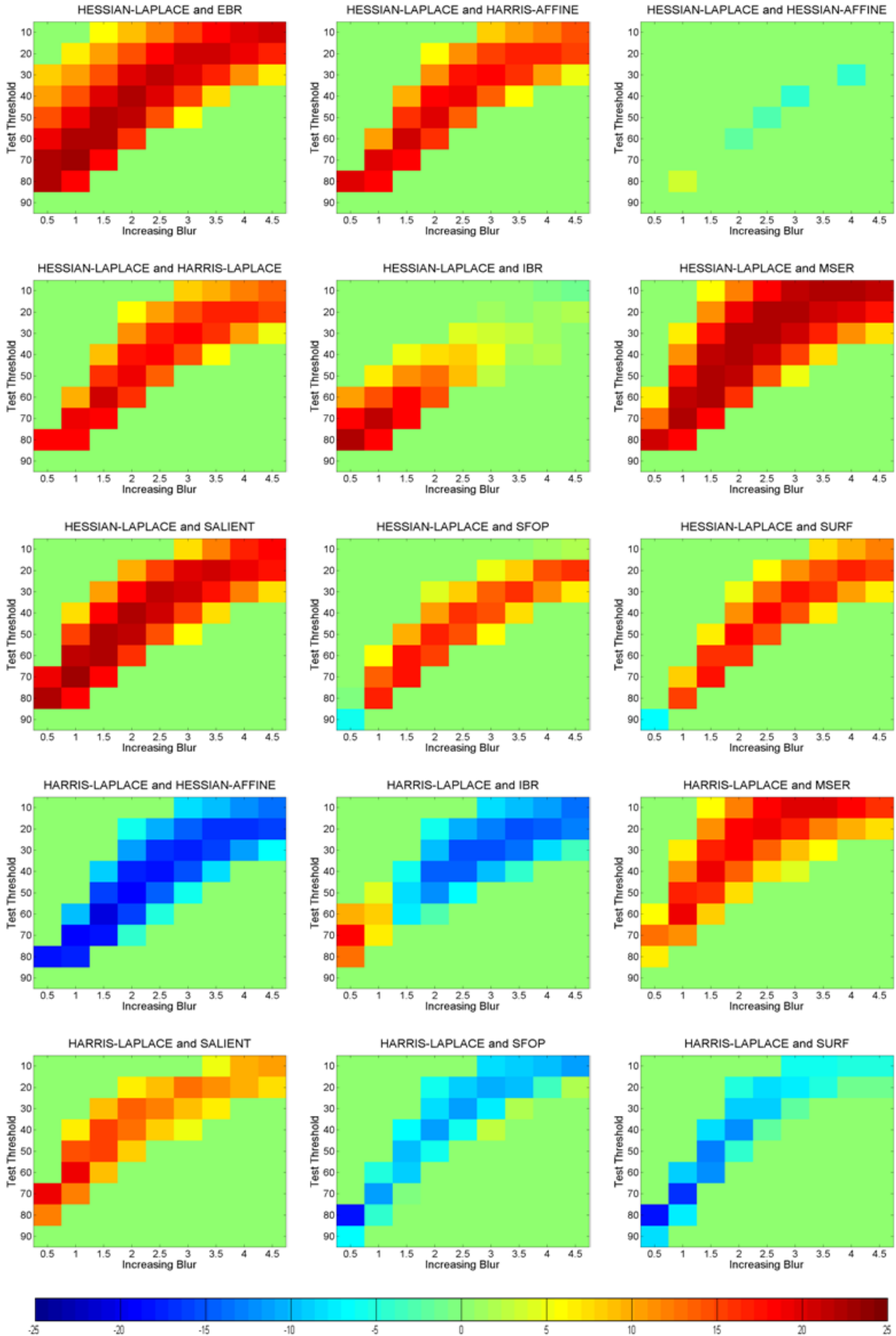

**Figure 14.** Results of blur changes for EBR and Harris-Affine with the other detectors showing Z-scores obtained; positive Z-scores indicate that the first detector is better than the second whereas the negative values show the converse.

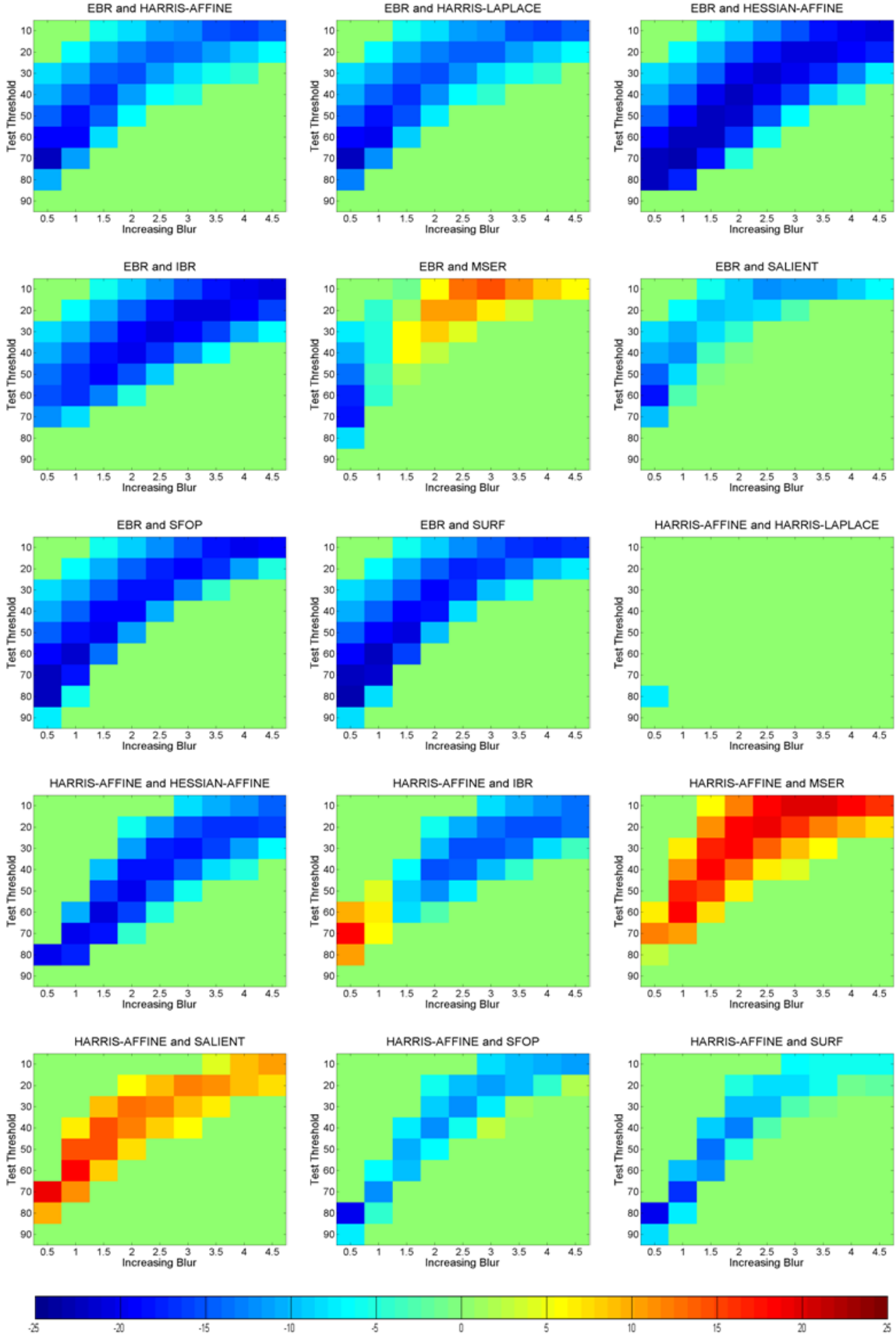

**Figure 15.** Results of blur changes for Hessian-Affine, IBR, MSER, SFOP and SURF with other detectors showing Z-scores; positive Z-scores show that the first detector is better than the second whereas negative values show the converse.

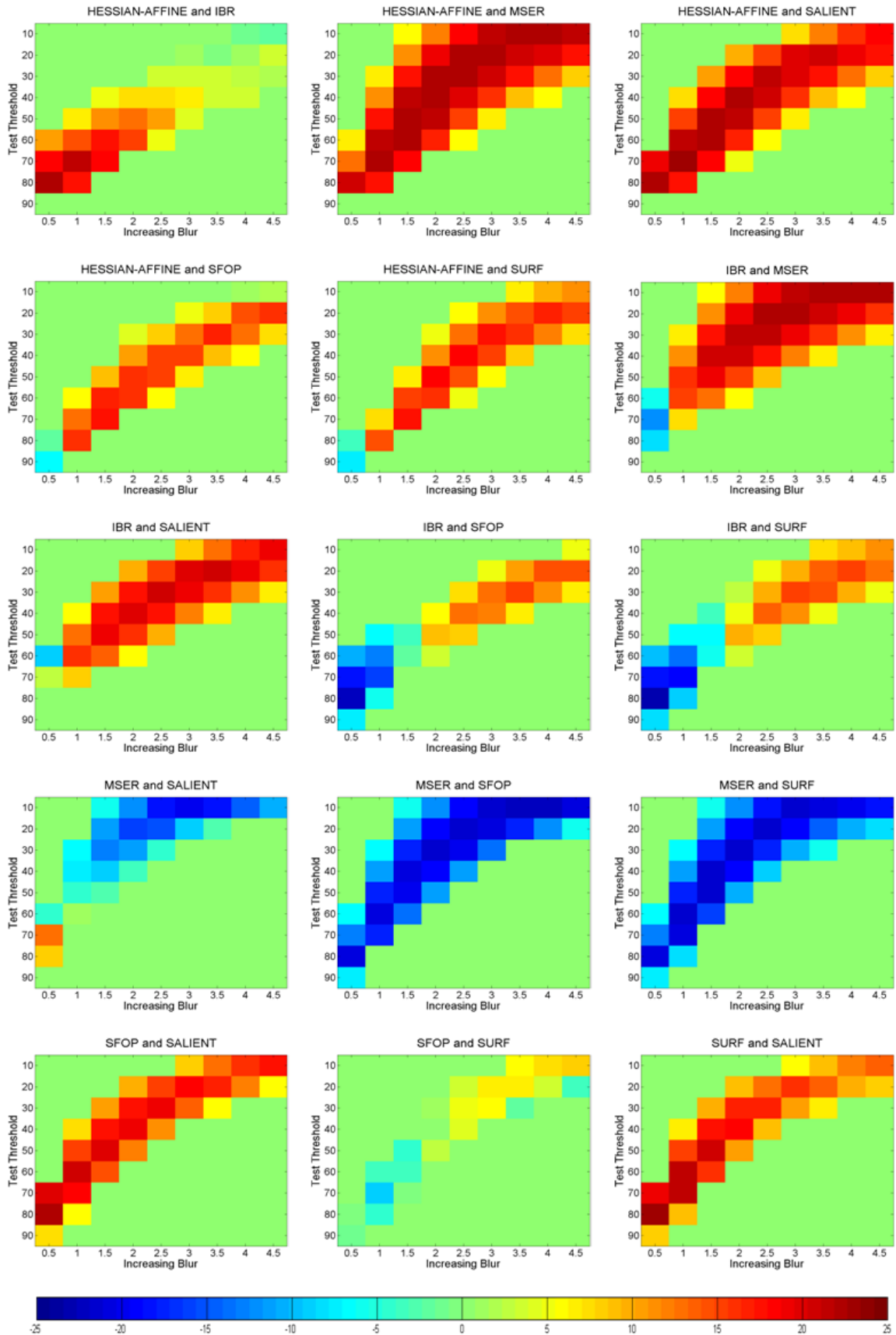

**Figure 16.** Results of uniform light changes for Harris-Laplace and Hessian-Laplace with other detectors showing Z-scores obtained; positive Z-scores indicate that the first detector is better than the second whereas negative values show the converse.

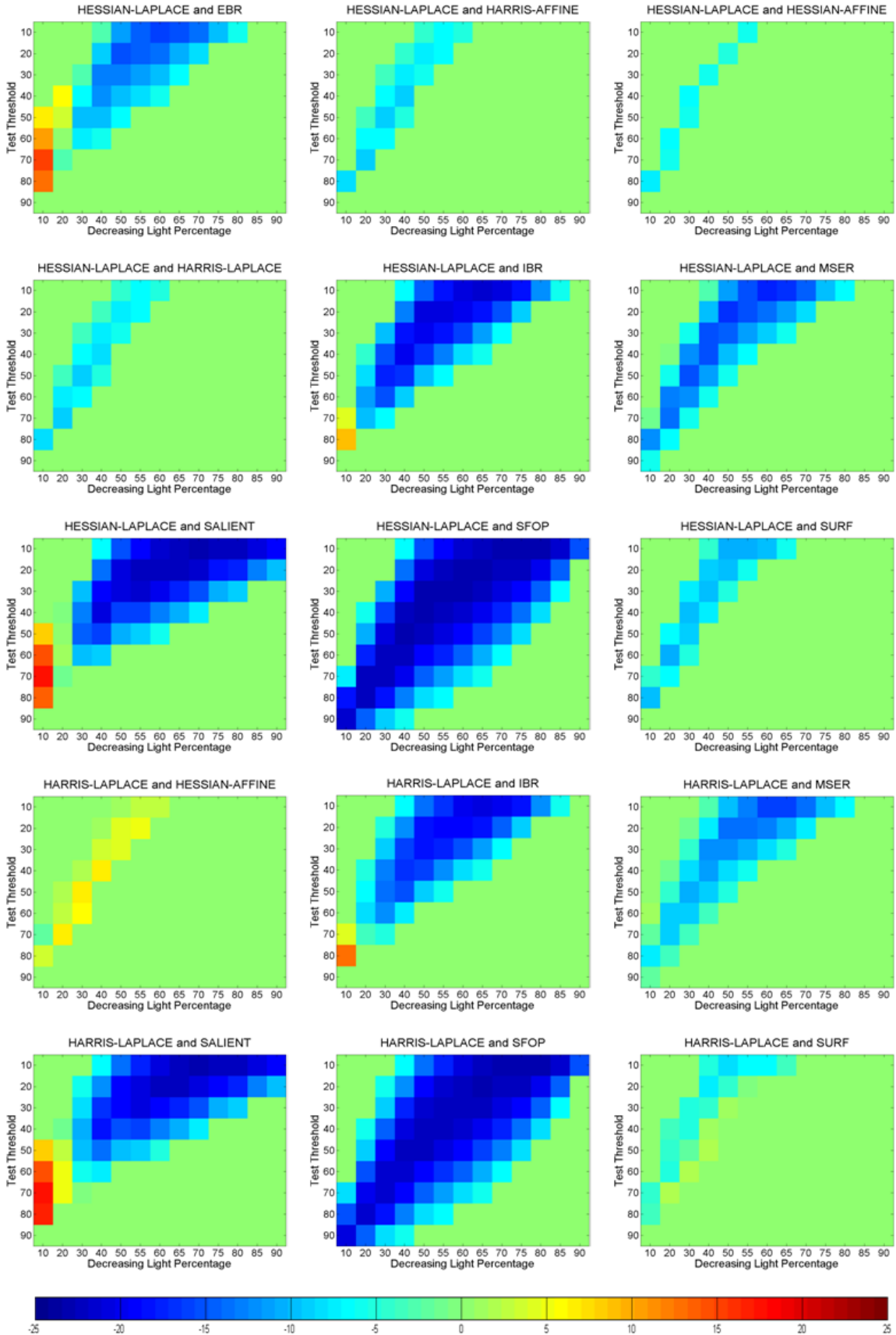

**Figure 17.** Results of uniform light changes for EBR and Harris-Affine with the other detectors showing Z-scores; positive Z-scores indicate that the first detector is better than the second whereas the negative values show the converse.

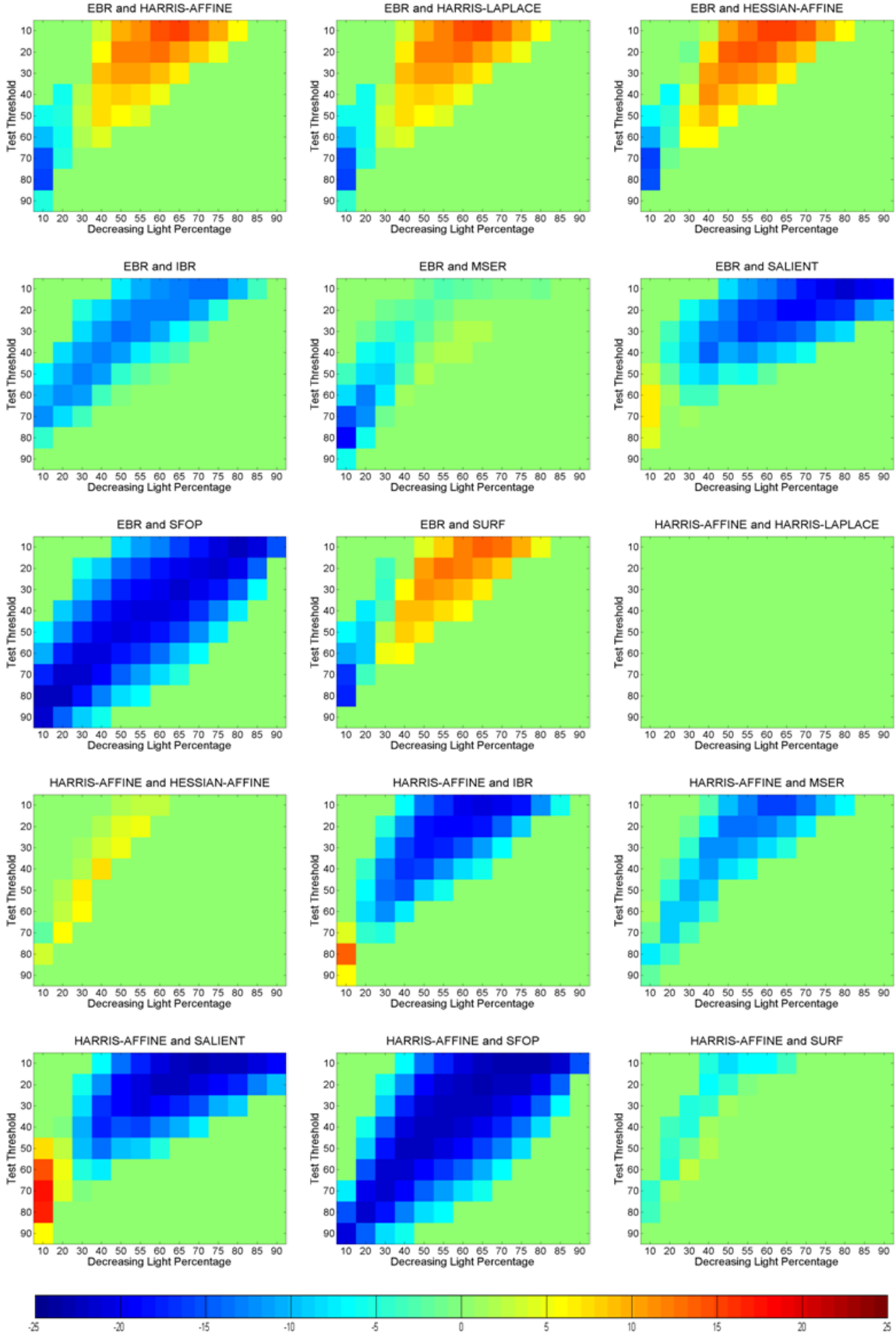

**Figure 18.** Results of uniform light changes for Hessian-Affine, IBR, MSER, SFOP and SURF with other detectors showing Z-scores; positive Z-scores show that the first detector is better than the second whereas negative values show the converse.

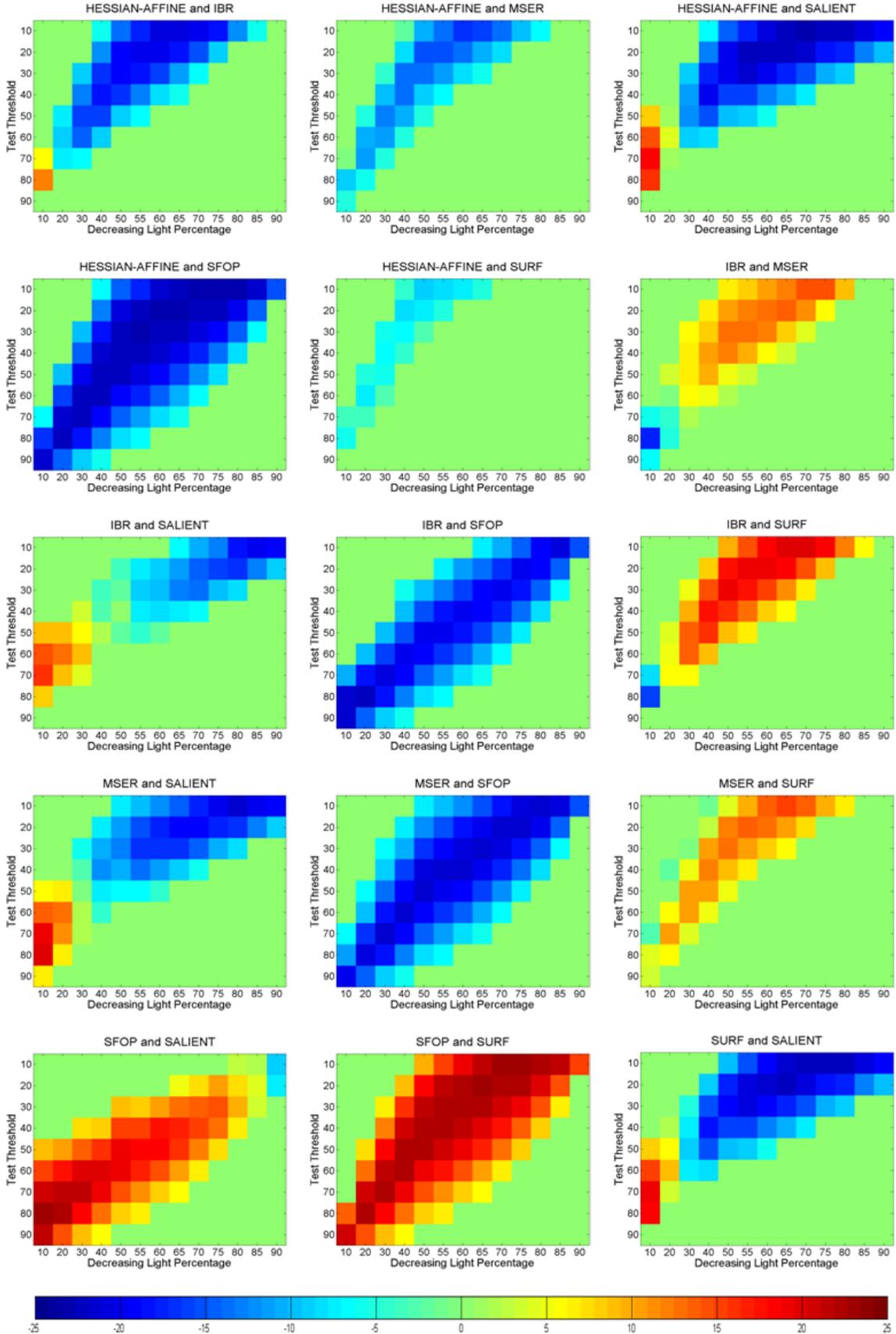

## 7. Conclusions

For designing reliable and more effective vision systems, this paper has presented a generic framework based on the improved repeatability measure [5]. The framework has two important steps: the first one is the estimation of the upper and lower bounds of performance for a given feature detector under a specific image transformation in order to segment detector performance into operating and guarantee regions; and the second step is the identification of statistically significant performance differences between detectors as a function of the amount of image transformation. To that end, the paper has introduced a variant of McNemar's test to find statistically significant performance differences. It has demonstrated the utility of the proposed framework by establishing operating and guarantee regions for several state-of-the art detectors and has identified statistical performance differences between them under JPEG compression, uniform light changes and blurring. Results are obtained by utilizing newly acquired, large image database with 539 different scenes for JPEG compression, blur and uniform illumination changes. These detailed results provide novel insights into the strengths and weaknesses of the detectors from a vision system design perspective. The results largely contradict the previous findings and provide new performance scores for the popular feature detectors under various image transformations. These performance curves are more consistent with what experienced vision researchers expect and encounter.

### Acknowledgments

This work was supported by the UK EPSRC Grant EP/K004638/1.